\documentclass[conference]{IEEEtran}
\IEEEoverridecommandlockouts
\usepackage{amsmath,amssymb,amsfonts}
\usepackage{algorithmic}
\usepackage{graphicx}
\usepackage{textcomp}
\usepackage{xcolor}
\usepackage{comment}
\usepackage[utf8]{inputenc}
\usepackage{color,soul}

\def\BibTeX{{\rm B\kern-.05em{\sc i\kern-.025em b}\kern-.08em
    T\kern-.1667em\lower.7ex\hbox{E}\kern-.125emX}}
\makeatletter
\def\@ConfName{Conferences Name}
\newcommand{\ConfName}[1]{\gdef\@ConfName{\small\sffamily#1}}

\def\@ConfAcronym{Conferences Acronym}
\newcommand{\ConfAcronym}[1]{\gdef\@ConfAcronym{\small\sffamily#1}}

\def\@ConfDate{Conferences Date}
\newcommand{\ConfDate}[1]{\gdef\@ConfDate{\small\sffamily#1}}

\def\@ConfCity{Conferences City}
\newcommand{\ConfCity}[1]{\gdef\@ConfCity{\small\sffamily#1}}

\def\@PaperNo{Paper Number}
\newcommand{\PaperNo}[1]{\gdef\@PaperNo{\large\sffamily#1}}

\def\@SetAffiliation{}
\newcommand{\SetAffiliation}[1]{\gdef\@SetAffiliation{\small#1}}

\def\@SetAuthors{}
\newcommand{\SetAuthors}[1]{\gdef\@SetAuthors{#1}}

\def\@Sponser{}
\newcommand{\Sponser}[1]{\gdef\@Sponser{#1}}

\def\@maketitle{%
	\newpage
	\null
	\vskip 2em%
	\begin{flushright}
		{\bfseries\sffamily \@ConfName\\[3pt]
		\@ConfAcronym\\
		\@ConfDate,
		\@ConfCity}\\
		\vspace*{1.0cm}
	{\bfseries\sffamily\Large	\@PaperNo}\\
	\vspace*{0.5cm}
	\end{flushright}
	\begin{center}%
		\let \footnote \thanks
		{\bfseries\LARGE\sffamily \@title \par}%
		\vskip 1.5em%
		{
			\lineskip .5em%
			\begin{tabular}[t]{c}%
				{\@SetAuthors}
			\end{tabular}\par}
		\vskip 0.5em%
	\@SetAffiliation
		\vskip 1em%
		{\large \@date}%
		\thanks{\noindent\rule{0.3\textwidth}{0.5pt}}
		\thanks{\@Sponser}
	\end{center}%
	\par
	\vskip 1.5em}

\makeatother
\renewcommand\IEEEkeywordsname{Keywords-component} 
\usepackage{footnote}
\renewcommand{\figurename}{Figure.}
\renewcommand{\tablename}{TABLE.}
\begin{document}
\title{Automated Fabrication of Magnetic Soft Microrobots }

\SetAuthors{
	Kaitlyn Clancy$^1$,
	Siwen Xie$^2$,
	Griffin Smith$^{1}$,
	Onaizah Onaizah$^{1,2*}$
	}

\SetAffiliation{
	$^1$School of Biomedical Engineering, McMaster University, Hamilton, Canada\\
	$^2$Department of Computing and Software, McMaster University, Hamilton, Canada\\
	$^*$E-mail: clanck2@mcmaster.ca, xies42@mcmaster.ca, smithg12@mcmaster.ca, onaizaho@mcmaster.ca
	}

\ConfName{Proceedings of the Canadian Society for Mechanical Engineering International Congress\\
          32nd Annual Conference of the Computational Fluid Dynamics Society of Canada\\
          Canadian Society of Rheology Symposium}
\ConfAcronym{CSME/CFDSC/CSR 2025}
\ConfDate{May 25--28, 2025} 
\ConfCity{Montréal, Québec, Canada} 
\PaperNo{NUMBER}

\Sponser{This work is supported by the Natural Sciences and Engineering Council of Canada (NSERC) through their Discovery Grant RGPIN-2024-06700, Canada Foundation for Innovation-John R. Evans Leader Fund through grant 44508 and the McMaster Entrepreneurship Academy.}

\date{May 25, 2025}

\maketitle

\begin{abstract}
\normalfont
The advent of 3D printing has revolutionized many industries and has had similar improvements for soft robots. However, many challenges persist for these functional devices. Magnetic soft robots require the addition of magnetic particles that must be correctly oriented. There is a significant gap in the automated fabrication of 3D geometric structures with 3D magnetization direction. A fully automated 3D printer was designed to improve accuracy, speed, and reproducibility. This design was able to achieve a circular spot size (voxels) of 1.6mm in diameter. An updated optical system can improve the resolution to a square spot size of 50$\mu$m by 50$\mu$m. The new system achieves higher resolution designs as shown through magneto-mechanical simulations. Various microrobots including 'worm', 'gripper' and 'zipper' designs are evaluated with the new spot size.
\end{abstract}

\vspace*{0.5em}
\begin{IEEEkeywords}
magnetics, soft microrobots, smart materials, optics, automation, fabrication
\end{IEEEkeywords}

\section{INTRODUCTION}
Microrobots are small end effectors that are capable of accessing small constrained environments and performing previously infeasible operations. As a result, they have gained interest in the scientific community because of their versatility in a wide range of applications. These highly functional materials have the potential to be used in various industries, such as biomedical, aerospace, electrical, and environmental applications \cite{KHALID2024112718}. They have demonstrated promising uses in targeted drug delivery \cite{doi:10.1021/acs.chemrev.1c00481}, minimally invasive procedures \cite{doi:10.1089/soro.2018.0136, article}, biosensing \cite{mi14061099}, and tissue engineering \cite{article2}.

Soft magnetic microrobots, a subset of robotic technology, are composed of magnetic microparticles embedded in a soft polymer matrix. These microparticles need to be oriented in specific directions to achieve the desired output motion in an externally applied magnetic field. As a result, these robots can achieve high degrees of freedom \cite{6392862}, as they can bend, twist, and flex in multiple directions, enabling complex structural formation and intricate task performance \cite{article3}. Magnetic actuation is used to drive translational and rotational motions to induce structural changes such as compression and expansion. These deformations support the actions of grabbing, rolling, swimming, and crawling, which are used to perform specific tasks such as drug delivery or wound patching \cite{doi:10.1021/acs.chemrev.1c00481, article3, article4}. 

Despite their vast potential, current fabrication technologies have limited capabilities. The production of intricate small-scale magnetic soft robots is a complicated tedious process that often requires days of manual labour \cite{PMID:37287461}. Fully automated fabrication systems are limited to the printing of two-dimensional (2D) geometric structures with three-dimensional (3D) magnetization directions or 3D geometric structures with 2D magnetization directions. These magnetization directions are determined by the orientation of the magnetic particles embedded in the polymer matrix, with 2D magnetization directions generally confined to planar orientations. To enable more complex designs and functionalities, it is essential to develop systems capable of producing 3D geometries with fully programmable 3D magnetization directions. In addition, automating the process enhances accuracy, increases production speed, and improves reproducibility, which are all vital for clinical applications.

\section{Background}

\subsection{Materials}

To fabricate these soft structures, numerous additive manufacturing methods use a wide variety of materials. Some use specific resins containing monomers for photopolymerization by light energy, whereas others are prepolymerized, for example thermoplastics such as polylactic acid (PLA) used in conventional 3D extrusion printing \cite{article2, PMID:37287461}. Advances in material science have led to the development of smart materials that respond to external stimuli such as heat, chemical gradients, pH, or magnetic fields \cite{KHALID2024112718}. Incorporating these stimuli responsive materials into the manufacturing and 3D printing process has led to the emergence of the term 4D printing \cite{mi14061099}. 

\subsection{Magnetic Actuation}
Developing a fully automated fabrication method for the production of soft magnetic robots requires an understanding of how magnetic materials are functionalized. When an external magnetic field is applied to actuate these robots, the embedded magnetic microparticles reorient to align with the external field. Therefore, the orientation of the particles determines the magnetization direction and is critical to the behaviour of the robots. Equation 1 shows the proportionality of the magnetic field (\textbf{H} measured in A/m) to the magnetic flux density (\textbf{B} measured in T) given the permeability ($\mu$ measured in $N/A^2$, H$/$m, or Tm$/$A) of the material. 

\begin{equation}
    |\textbf{B}| = \mu|\textbf{H}| 
    \label{eq:required_steps1}
\end{equation}

This allows us to model the magnetic forces and torques caused by the magnetization direction of a soft structure.

\begin{equation}
   \textbf{F} = \mu_0v(\textbf{M}\cdot \nabla)\textbf{H} 
    \label{eq:required_steps2}
\end{equation} 
\begin{equation}
    \textbf{$\tau$} = \mu_0v\textbf{M}\times\textbf{H}
    \label{eq:required_steps3}
\end{equation} 

Applying Maxwell’s equations to materials without a flowing current, the force is simplified to:

\begin{equation}
    \textbf{F} = \mu_0v [\textbf{M} \cdot \frac{\partial \textbf{H}}{\partial x} ,\textbf{M} \cdot \frac{\partial \textbf{H}}{\partial y}, \textbf{M} \cdot \frac{\partial \textbf{H}}{\partial z}]^T
    \label{eq:required_steps4}
\end{equation} 

These equations enable the prediction and modeling of the actuation of magnetic robots. Three common types of magnetic fields are used to achieve a wide range of motion: field gradients, uniform fields, and rotational fields. In field gradients, translational movement will occur in the direction of increasing magnitude \cite{article5}. Rotational magnetic fields generate a constant torque, which is maximum when the magnetic particles are oriented perpendicular to the magnetic field \cite{article5}. Fig. 1 illustrates magnetic actuation under a non-uniform field. By applying a rotating magnetic field, motion such as rolling and propulsion in liquid environments are achieved \cite{article5}. Uniform external magnetic fields are used to isolate rotational motions at a specific point. This allows for precise and accurate control over a manipulated structure. 

\begin{figure}[h]
    \centering
    \includegraphics[width=0.6\linewidth]{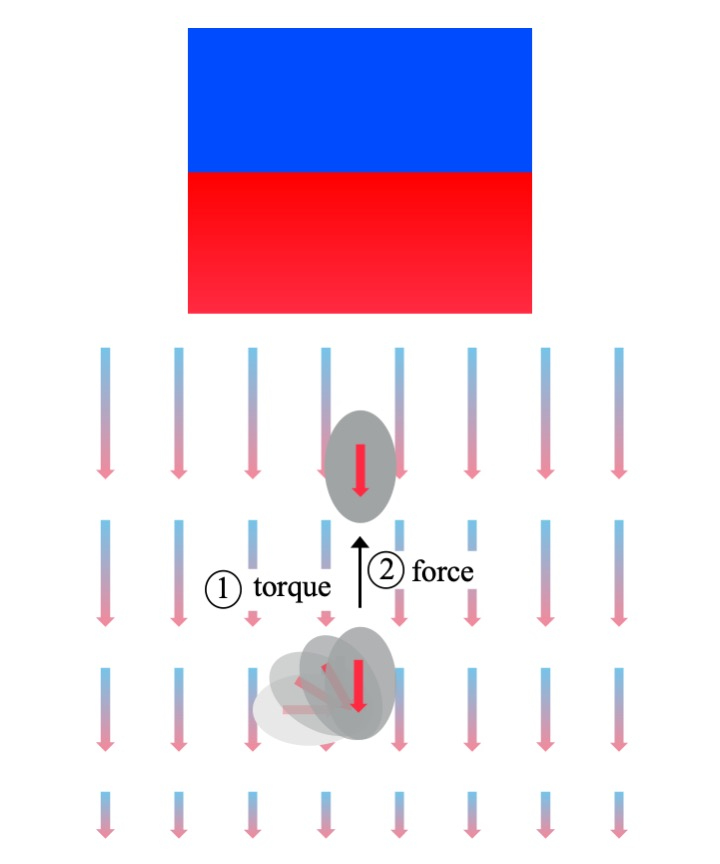} 
    \caption{Magnetic torque and force illustration in a non-uniform magnetic field generated from an external permanent magnet.}
    \label{fig:Fig1}
\end{figure}

\section{Design}
\subsection{Previous Design}
\noindent \textbf{\textit{Printer Characterization:}} Expanding on previous light-based fabrication systems \cite{article6, article7, https://doi.org/10.1002/adfm.201802110}, a stereolithography (SLA) based 3D printer was successfully designed to encode 3D magnetization directions into 3D printed geometric structures and this process was fully automated \cite{https://doi.org/10.1002/adrr.202400005}. The printer uses a baseplate to move the composite resin well through 2D space, and the third dimension is achieved by a build plate that lowers into the well. A three degree-of-freedom (DOF) rotating permanent magnet reorients the magnetic microparticles along the direction of the external magnetic field. The ultraviolet (UV) light emitting diode (LED) then programs this orientation into a voxel by curing a local region of resin. The stepwise translate, reorient, and cure process repeats voxel by voxel, then layer by layer until the user-designed magnetic soft robot is produced.

\subsection{Updated Design} 

In this study, the UV LED and relevant printed circuit board (PCB) are updated to achieve higher resolution voxels to accurately print soft microrobots. In addition, these will be uniformly cured in the shape of a square. The new components included in the updated optical system are a compact laser diode/temperature controller with a mount, a pigtailed laser diode with a collimator, a near ultraviolet (NUV) flat top beam shaper, and a fused silica broadband dielectric mirror, which are described in detail below. The updated optical system's layout can be seen in Fig. 2. The laser diode is emitted through the beam shaper, reflected off the mirror, and through a lens to focus the spot size onto the build plate.

\begin{figure} [h]
    \centering
    \includegraphics[width=1\linewidth]{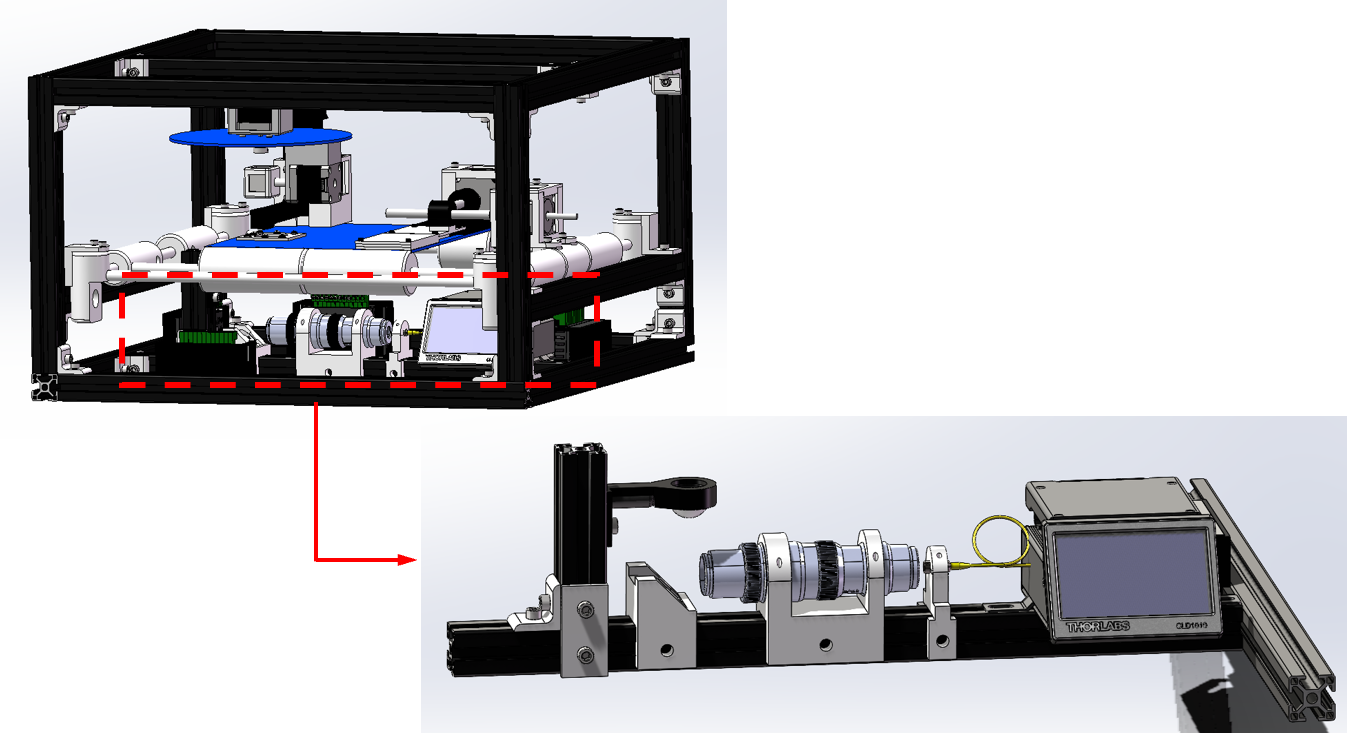}
    \caption{Assembly of the updated optical system for the 3D printer.}
    \label{fig:Fig2}
\end{figure}

In general, the updated optical system can significantly improve design performance and fabrication precision. The integration of the NUV flat top beam shaper and the collimator with the pigtailed laser diode to the system achieves uniform beam energy distribution to avoid tapered curing at the voxel edges. These components also optimize the design to reduce the spot size from a circle that is 1.6mm in diameter to a square that is 50$\mu$m by 50$\mu$m for higher resolution and fine microstructure printing capabilities. In addition, the selected LP405C1 laser has a power range of 30-40mW and a wavelength of 405nm, which integrates well with the previous system's PCB and material parameters, ensuring curing efficiency and performance. Using the compact laser diode/temperature controller with a mount, the new system has better control of the temperature and current to the laser diode, preventing overload and overheating, and ensuring the stability and safety of the equipment by using automated power (on/off) management.
 
\subsubsection{{Compact Laser Diode/Temperature Controller (LD/TEC) with Mount}}
The CLD1010LP Compact Laser Diode/Temperature Controller was selected to ensure accurate and stable control of the laser diode. This driver has the ability to deliver constant current or power. It also has the ability to maintain the temperature of the laser, preventing overheating of the system.  This is crucial to improve the performance of the optical system; slight fluctuations of these aspects may affect the beam's shaping effect and the printing quality of the microstructure. Although this driver has a touch panel for easy use, it can be accessed remotely with a universal serial bus (USB) or connected through its external modulation feature. The modulation can go up to 300kHz for this particular model. To complement the laser control, it has a mount embedded into the system providing the main support and stabilization to the laser. This mount is designed with a Clamshell to effectively prevent physical damage and reduce the thermal gradient on the surface of the laser diode, which improves the overall heat dissipation efficiency. In addition, its cooling module is in close contact with the fiber tail housing, ensuring temperature stability throughout operation, and sustaining the laser's output power and wavelength. 

\subsubsection{Pigtailed Laser Diode with Collimator}
The pigtailed laser diode with collimator (LP405C1) provides a single-mode fiber-coupled laser diode with a wavelength of 405nm, a typical optical output power of 30mW, and a sealed argon-filled collimator. The included collimator reduces the contamination of fine particles, ensuring stability in long-term operation. Furthermore, the laser is equipped with fiber caps at the input and output ends, which effectively reduces the optical power density of the interface and prevents common air-to-fiber material deposition, thereby maintaining long-term optical performance. The collimated output of the laser has an 8mm spot diameter and uses a high-precision alignment process to maximize the coupling of the laser emission to the single-mode fiber. This improves the beam shape and provides an ideal input light source for the subsequent beam shaper.

\subsubsection{Near Ultraviolet (NUV) Flat Top Beam Shaper}
The flat top beam shaper converts the Gaussian beam intensity into a uniform intensity distribution, significantly improving the accuracy and consistency of resin curing. Based on the refractive field mapping design, the beam shaper uses a combination of two optical elements to redistribute energy. The first element introduces spherical aberration and the second eliminates the aberration through compensation, ultimately forming a collimated beam with a flat wavefront, low divergence, and high-intensity uniformity at the output end. Fig. 3 demonstrates the ability of the beam shaper to convert a light source between 335nm to 560nm into uniform energy distributions while maintaining its stability over long distances (150mm to 200mm). This beam shaper matches the input beam from the selected laser diode and provides a stable output while maintaining a low wavefront error. This is crucial for uniform beam curing, ultimately solving the delamination problem in the previous design. It also produces higher-resolution microstructures and provides well-defined square voxels, more closely matching the user-defined designs on COMSOL. 

\begin{figure} [h]
     \centering
     \includegraphics[width=1\linewidth]{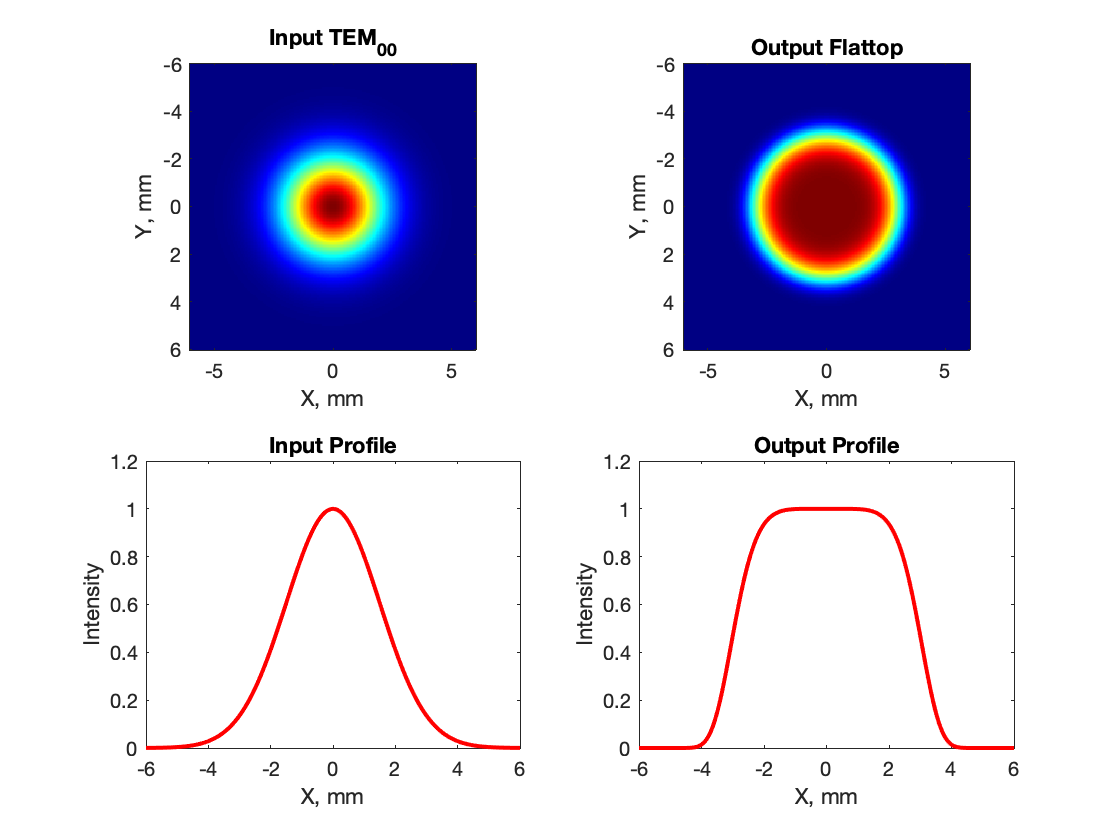}
     \caption{Improvements to output intensity distribution with the addition of a beam shaper.}
     \label{fig:Fig3}
\end{figure}

\subsubsection{Fused Silica Broadband Dielectric Mirror}
Once the beam has been converted to a uniform square output distribution, a 1" diameter fused silica broadband dielectric mirror alters the beam's trajectory by 90°. This mirror uses a high-quality fused silica substrate combined with a broadband dielectric coating for excellent reflection performance. It can provide an average reflectivity of more than 99\% for S- and P-polarized light in the spectral range of 400nm to 750nm. In practical applications, the mirror accurately reflects and guides the beam towards the buildplate, enabling efficient use of space within the optical setup. 

\section{Fabrication}
\subsection*{Solidworks Designs}
To integrate all of the components together, computer aided design software (SolidWorks 2024) was used to model the entire design. Specifically, the support structures, seen in Fig. 4, were designed to place all of the new components on the same axis in a stable structure, ensuring precise alignment during operation and minimizing errors caused by offsets or vibrations. 

Each support is built to hold a single component. Specifically, a mounting seat with an angle of 45° was designed for the fused silica broadband dielectric mirror to accurately alter the beam trajectory by 90°. The NUV flat top beam shaper utilized a stable double-clamping fixed position and precise alignment grooves to facilitate beam homogenization, while still allowing for changes to the shapers output parameters. Furthermore, a similar clamping structure was used for the pigtailed laser diode to tightly fix the fiber tail and prevent vibrational deviations. This holds the output end of the laser diode in place. 

\begin{figure} [h]
    \centering
    \includegraphics[width=0.9\linewidth]{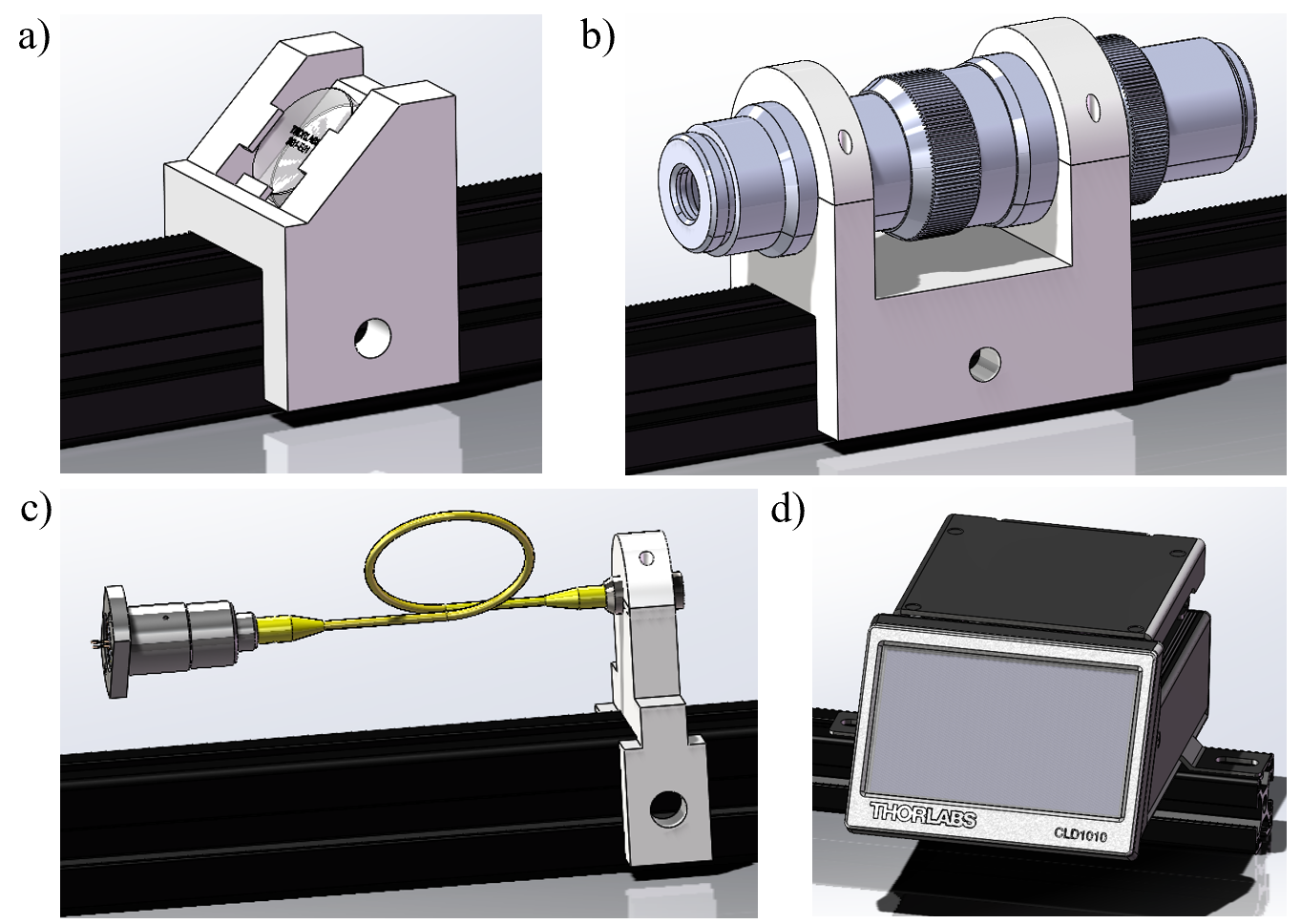}
    \caption{Housing for the new optical system: a) Fused silica broadband dielectric mirror mounting seat. b) NUV flat top beam shaper double clamping case. c) Pigtailed laser diode fiber tail clamp. d) LD/TEC mount.}
    \label{fig:Fig4}
\end{figure}

\section{SIMULATIONS}
\subsection{Optical Simulation}
A simulation of the integration of the new components was created to optimize the spot size of the laser beam at the cure site. This entire process was designed in COMSOL Multiphysics 6.2 using the Geometrical Optics package, along with a Ray Tracing study. To begin, a point source for the pigtailed laser diode was created and set to emit rays as a conical direction vector to mimic the beams' trajectory. The NUV flat top beam shaper was simulated as a 6mm by 6mm by 139mm square column placed in front of the laser diode, converting the laser's divergence pathways into parallel rays. These rays were then reflected off the silica broadband dielectric mirror which was simulated at a 45° angle and assumed to have no ray absorption. Lastly, the rays were focused using a N-BK7 plano-convex lens for ray convergence, ultimately reducing the spot size. 
From the simulation, the convergence point was determined to be 19.8mm from the top of the focusing lens, as shown in Fig. 5 a). The convergence point is the distance of the lens under the resin well. Overall, the beam spot size was found to be a square as shown in Fig. 5 b) and c), which was measured to be 50$\mu$m by 50$\mu$m at the convergence point. Based on the history of simulations conducted on the previous printer's optical system, the actual spot size will differ slightly from the predicted value.

\begin{figure} [h]
        \centering
        \includegraphics[width=1\linewidth]{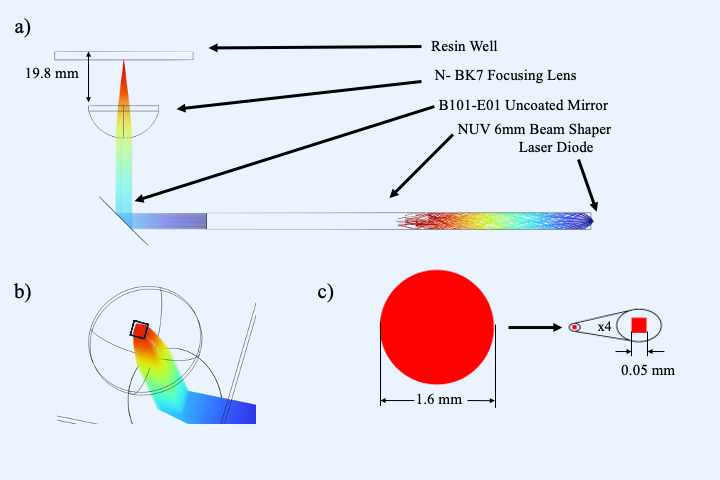}
        
        \caption{COMSOL optics simulation. a) Optics set up with laser beam simulation. b) Confirmed voxel output shape. c) Voxel shape and size evolution.}
    \label{fig:Fig5}
\end{figure}

\subsection{Microrobot Design Simulations}
Once the new spot size was generated and the square output confirmed, various test designs were developed to evaluate the printer's capabilities. First, a simple "worm" structure with uniform magnetization was designed for beam bending analysis. The encoded magnetization was simulated under a constant magnetic field to observe deflection, as shown in Fig. 6. The observed deflection is shown for a 4mT flux density given an elastic modulus of 4.6MPa, a Poisson's ratio of 0.49, and a relative permeability of 2.75MPa. 

\begin{figure} [h]
    \centering
    \includegraphics[width=1\linewidth]{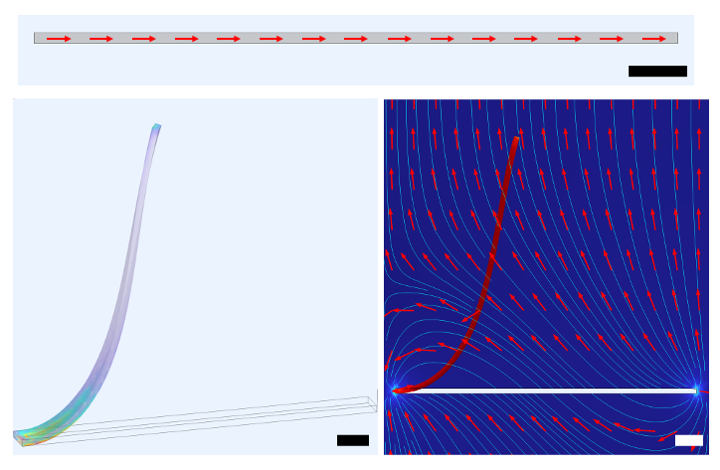}
    \caption{"Worm" design. a) Magnetization. b) Bending simulation showing start and end position. c) Bending simulation showing applied magnetic field. Each scale bar represents 50$\mu$m.}
    \label{fig:Fig6}
\end{figure}

In addition to the worm, the "gripper" illustrated in Fig. 7 is used to demonstrate the printer's ability to 3D magnetize structures with 3D geometries. This design is capable of grasping and transporting objects. 

\begin{figure} [h]
    \centering
   \includegraphics[width=1\linewidth]{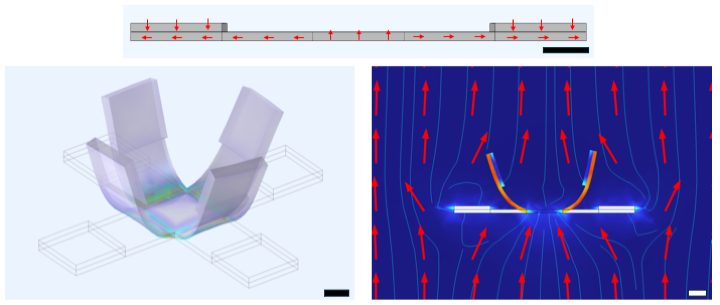}
   \caption{"Gripper" design. a) Magnetization. b) Bending simulation showing start and end position. c) Bending simulation showing applied magnetic field. Each scale bar represents 25$\mu$m.}
    \label{fig:Fig7}
\end{figure}

Lastly, given the new voxel shape of the printer, a "zipper" model was designed to test the printer's capabilities to sufficiently connect the corner of each voxel, as shown in Fig. 8. If the voxels of such a structure do not consistently adhere to one another, then a slight overlap of voxels would be tested. This is shown in Fig. 8 b). The overlap of voxels in past models would result in partial curing and delamination. However, the new beam shaper has the ability to create a uniform energy distribution across the spot size and should remove these past limitations. 

\begin{figure} [h]
    \centering
    \includegraphics[width=1\linewidth]{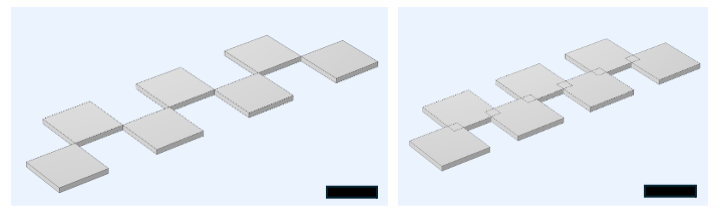}
    \caption{"Zipper" design. a) Direct corner connection. b) Corner overlap. Each scale bar represents 50$\mu$m.}
    \label{fig:Fig8}
\end{figure}

Each of the above designs can be used to test the accuracy, curing distribution, and repeatability of the new voxel size.

\section{ALGORITHM}
An aspect that enables repeatability is the complete automation of the printing process. To reduce the need for manual steps that are typically required, an algorithm was adopted from previous iterations of the 3D printer. The code obtains important magnetization and geometric data from COMSOL designs and interprets the positions of each geometry to determine the minimal path to cure and magnetize each voxel. This eliminates the need for manual assembly and reduces the time and cost of fabrication. The overall workflow followed is outlined in Fig. 9. 

\begin{figure} [h]
    \centering
    \includegraphics[width=1\linewidth]{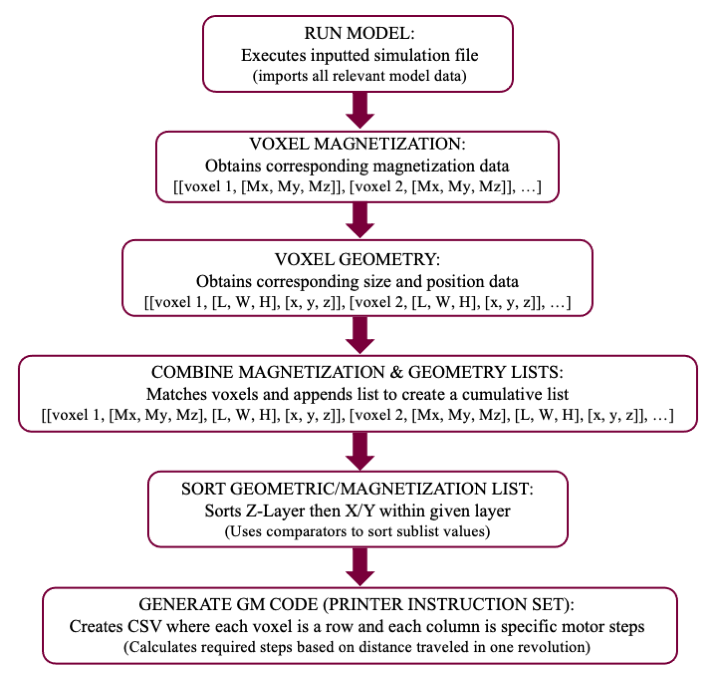}
    \caption{The steps for the printer automation algorithm.} 
    \label{fig:Fig9}
\end{figure}
 
This process begins after the user designs the desired models. From here the code extracts two main pieces of data from the COMSOL Multiphysics Software to characterize each voxel for fabrication. These unique building blocks have a domain ID to identify them and allow for position-based sorting to be automated for optimal paths during fabrication. The two main arrays obtained from COMSOL are the magnetization and geometry datasets. The magnetization data provides the ID\#, Mx, My, and Mz magnetization directions, whereas the geometry includes the ID\#, length (L), width (W), and height (H) of each building block as well as their corresponding position in x, y, z coordinates. These two datasets are then combined according to the same voxel ID so that the full range of information can be accessed for each one. To optimize the path, the voxels are first sorted based on their z position, prioritizing those highest first because they are cured from the top-down. Within each of these geometric layers, the voxels are organized based on those closest to the origin by calculating their hypotenuse. Finally, the G-code is generated to create an instruction set for the 3D printer. It is important to note each distance is converted to the rotational steps the motor needs to take based on their resolution. 

The magnetization must also be converted from Cartesian coordinates to spherical for proper motor control. The motor steps required to achieve the desired positions are outlined in Eqns. 5 to 7. The translational distance required for each motor (x, y, z) is converted to number of revolutions by following Eqn. 5. 

\begin{equation}
    \text{requiredTransRevs} = \frac{\text{motorPosition}}{\text{distancePerRev}}
    \label{eq:required_revs}
\end{equation}

To transfer between voxel locations, the change between revolution sets is calculated using Eqn. 6. 

\begin{equation}
    \text{requiredTransSteps} = {\text{stepsPerRev}}\times{\text{requiredTransRevs}}
    \label{eq:required_steps}
\end{equation}

The magnetization motors incorporate the spherical angle coordinates with one motor for the azimuth angle and one for the inclination angle individually. The equation used to calculate the number of motor steps based on the desired motor angle and degree's in one step is shown in Eqn. 7. The motorAngle and motorPosition variables used for the motors are calculated based on the desired position subtracted from the current position as shown for the translational position changes. This data is all written into a G-code by the initial slicer algorithm and read by the Raspberry Pi code to generate motor movement.

\begin{equation}
    \text{requiredRevoSteps} = \frac{\text{motorAngle}}{\text{degreePerStep}}
    \label{eq:required_steps_magnetization}
\end{equation}   

 
\section{Future Work}
The implementation of the new optics system for the 3D printer requires several essential characterization tests. This is currently being tested by functionally integrating the 3D printed support structures shown in Fig. 2 and securely fixing the new components. 
The first test must determine the optimal curing time for the new laser. This can be achieved by altering the curing duration and evaluating the resulting cure and build quality. This is an essential parameter for the automation process, ensuring optimal print quality while minimizing production time. 
Once the optimal cure time is established the geometry of a single voxel must be characterized. This will inform the design constraints imposed when creating new microrobot structures in COMSOL. The smallest spot size that can be cured by the new laser was simulated and shown in Fig. 5 but needs to be experimentally tested and validated. Finally, once the 3D printer is effectively printing single voxels, the designs above will be printed to assess the overall build and curing quality at different resin to magnetic particle ratios. These results will then be used to refine COMSOL simulation parameters for more accurate shape morphing predictions. 

\section{Conclusion}
We simulated an updated optical system for a previously designed SLA-based 3D printer to achieve higher resolutions. By adopting the LP405C1 laser diode and introducing a beam shaper, this project has optimized the optical system to reduce the spot size from 1.6mm to about 50$\mu$m. The beam shaper ensures well-defined curing boundaries, while eliminating the tapered solidification problem at the edge of previous voxels, reducing the delamination phenomenon and significantly improving the overall stability of the printed structure. In addition, the integration of the external current drive system ensures precise control of the laser power, effectively preventing the laser diode from overheating, and ensuring the stability and consistency of the system's operation. Mechanically, the optical components are precisely aligned on the same axis through a modularized fixed board structure, ensuring the stability and consistency of the optical path. The magnetization data and geometric data of the voxels can be obtained through the interface with COMSOL. The code can be implemented by the Raspberry Pi to realize the automatic control of the entire printing process, which greatly improves the operation efficiency and system stability. For validation, this project carried out an optical and magneto-mechanical simulations of various soft magnetic robots. Specifically, this includes a variety of designs based on COMSOL, such as a “worm”, “gripper”, and “zipper” to test the system’s performance under different geometric complexities and motion patterns.

\bibliographystyle{ieeetr}
\bibliography{ref}

\end{document}